# Generating a Biometrically Unique and Realistic Iris Database

Jingxuan Zhang, Robert, J. Hart, Ziqian Bi, Shiaofen Fang and Susan Walsh.

*Abstract*— **The use of the iris as a biometric identifier has increased dramatically over the last 30 years, prompting privacy and security concerns about the use of iris images in research. It can be difficult to acquire iris image databases due to ethical concerns, and this can be a barrier for those performing biometrics research. In this paper, we describe and show how to create a database of realistic, biometrically unidentifiable colored iris images by training a diffusion model within an open-source diffusion framework. Not only were we able to verify that our model is capable of creating iris textures that are biometrically unique from the training data, but we were also able to verify that our model output creates a full distribution of realistic iris pigmentations. We highlight the fact that the utility of diffusion networks to achieve these criteria with relative ease, warrants additional research in its use within the context of iris database generation and presentation attack security.**

This work was supported in part by the U.S. Department of Justice under Grant 15PNIJ-23-GG-04206-RESS. (*Corresponding author: Susan Walsh*). Jingxuan Zhang and Robert J Hart contributed equally to this work and are co-first authors.
Jingxuan Zhang, Ziqian Bi and Shiaofen Fang are with the Luddy School of Informatics, Computing & Engineering, Indiana University Indianapolis. IN 46202 USA.
Susan Walsh and Robert J. Hart are with the Department of Biology, School of Science, Indiana University Indianapolis, IN 46202 USA. (email: walshsus@iu.edu).
Color versions of one or more of the figures in this article are available online at http://ieeexplore.ieee.org

## I. INTRODUCTION

The human iris, a colored and patterned ring-like structure situated between the pupil and sclera, possesses intricate textures formed by the interplay of blood vessels, muscles, collagen, and melanin. These unique patterns make the iris an ideal candidate for biometric identification.

In an 1888 lecture at the Royal Institution of Great Britain, Francis Galton first recognized the potential of iris patterns for identifying individuals [1]. However, his focus soon shifted to fingerprints, where he became a pioneer in identification research. Decades later, at the 41st annual meeting of the American Academy of Ophthalmology and Otolaryngology in 1936, Frank Burch proposed the use of iris patterns for identification [2,3]. Despite this early suggestion, it wasn't until 1986 that ophthalmologists Leonard Flom and Aran Safir filed the first patent for an iris identification system [4]. Their patent outlined a broad concept involving the illumination, photographing, storing, and comparison of irises, but it lacked the specificity required for practical implementation.

The breakthrough came in 1993 when John Daugman published a detailed description of an operational iris recognition system[5], earning a patent the following year[2]. Daugman's system was groundbreaking for two reasons: 1) it provided the technical specificity needed to realize iris-based biometric identification; 2) it offered empirical proof that no two irises are identical even between the two eyes of one person or between identical twins[6].

The foundational work of pioneers established the iris as a reliable biometric identifier. Building on this foundation, *Section A* explores Daugman's approach and its influence on traditional iris identification systems. *Section B* examines current challenges in iris biometric research, including non-ideal imaging conditions and security vulnerabilities. *Section C* highlights the scarcity of high-quality iris datasets and its impact on research progress. In *Section D*, we discuss the potential and risks of diffusion networks in generating synthetic iris images. Finally, *Section E* outlines our approach: leveraging diffusion models to create realistic, biometrically unique iris textures while addressing security concerns.

### A. Daugman's Approach and Traditional Iris Identification Systems

Daugman's approach to iris identification begins with capturing an image of a subject's eye. The iris is isolated by detecting its limbic and pupillary boundaries. Next, the iris is normalized through a transformation into a polar coordinate system of fixed dimensions. This normalization addresses resolution differences between images and reduces the size of high-resolution data, optimizing computational efficiency. Feature extraction is then performed using 2D complex Gabor filters, which analyze the texture patterns within the iris. Each response from these filters is quantized into binary values based on phase information, resulting in a compact iris code template. The similarity between iris templates is measured using the Hamming Distance (HD). If the HD value falls below a predefined threshold, the two irises are statistically dependent and considered a match [7].

An ideal HD threshold effectively distinguishes between authentic comparisons (images of the same iris) and imposter comparisons (images of different irises). Daugman's system demonstrated exceptional accuracy, achieving a false acceptance rate of 1 in 1,000,000 after more than 200 billion real-world comparisons, using an HD threshold of 0.317 [8].

Following the initial steps of image acquisition and iris segmentation, most traditional iris identification systems adhere to a core sequence inspired by Daugman:

1.) Iris normalization into a polar plane;
2.) Feature extraction using texture filters;
3.) Quantization of the filter response into a binary iris code template;

    4.) Calculation of similarity between iris code templates for criterion-based match determination.

In addition to this sequence, most modern iris systems incorporate three optimizations:

    1.) Near-infrared imaging to reduce specular reflections and improve texture detail.
    2.) Noise masking algorithms to handle obstructions like eyelashes, eyelids, and reflections.
    3.) Exclusion of masked iris code bits during similarity calculations.

These refinements, though not explicitly described in Daugman's original works, are evident in the real-world implementations of his system. Additionally, advancements like 1D Log-Gabor filters [13] and 2D Log-Gabor filters [9] have provided further accuracy improvements.

In summary, Daugman's methodology laid the foundation for modern iris identification systems. While the core principles remain unchanged, subsequent optimizations have significantly improved robustness and accuracy, enabling reliable performance across diverse real-world scenarios.

### B. The Atmosphere of Iris Biometric Research

The growth of iris biometrics research over the past three decades reflects its increasing importance in the realm of biometric identification. A rough measure of this growth can be seen through Google Scholar search results: between 1991 and 1995, only 4 publications referenced 'iris biometrics'. By 2001 to 2005, this number surged to 982, and between 2016 to 2021, it reached 1,870. These numbers underscore the rapid expansion and sustained interest in the field.

A significant portion of research focuses on improving performance when dealing with non-ideal iris images [32,33]. Non-ideal images arise from factors such as occlusions (eyelids, eyelashes, glasses), specular reflections, extreme angles, long capture distances, and iris deformations caused by injury or congenital disorders. The variability in surrounding eye regions poses challenges in extracting meaningful biometric features. Therefore, advancements in image acquisition and image preprocessing algorithms remain key areas of exploration.

Security is another critical concern in iris biometric systems. Presentation attacks, where spoofed or artificially generated irises are presented to fool the system, pose significant risks. Additionally, reverse engineering of iris code templates can generate images resembling the original iris [10]. This dual vulnerability raises both privacy concerns and security risks, making iris templates attractive for malicious actors.

Common countermeasures include liveness detection algorithms, which ensure the presented iris belongs to a living human. These algorithms are crucial in mitigating spoofing attempts and maintaining system integrity.

    The adoption of machine learning (ML) and artificial intelligence (AI) has further transformed the landscape. Neural networks and deep learning models are increasingly used as alternatives to traditional feature extraction and matching techniques. These models offer advantages in handling complex patterns, adapting to non-ideal images, and improving resistance to attacks.

In conclusion, iris biometric research continues to thrive, addressing challenges in image quality, system security, and algorithmic performance. The integration of AI and advanced algorithms represents a promising direction for overcoming existing limitations and expanding the capabilities of iris identification systems.

### C. The Role and Availibility of Iris Image Databases

Research, improvement, and innovation of iris identification systems inherently relies on images of iris images. However, increasing security and privacy concerns surrounding the iris make iris databases increasingly hard to come by. In 2019, [5] surveyed publicly available databases of real and synthetic irises. Only 81 of 158 iris databases purported to be public were actually available to them [5], and of the images they were able to acquire, only 8.8% of them were taken in the visible spectrum [5]. Further, [5] also points out that many available iris datasets are kept behind legal releases, some of which must be signed by legal representatives for the institutions of those requesting access.

The lack of available iris image datasets imposes barriers upon those wishing to conduct iris biometric research, limiting the number of papers, perspectives, and solutions in the field. As data limitations persist, generative models like diffusion networks offer a promising avenue for creating synthetic iris datasets.

### D. Advantages and Challenges of Diffusion Networks in Synthetic Iris Image Generation

In recent years, diffusion models, particularly Denoising Diffusion Probabilistic Models (DDPMs) [24], have emerged as a dominant force in the field of image synthesis, surpassing traditional generative adversarial networks (GANs) [32] in both image quality and generation stability. Diffusion models follow a step-by-step denoising process that progressively refines random noise into high-quality images. This iterative refinement results in superior image fidelity and intricate texture representation, making them ideal for generating the complex patterns and pigmentation of human irises.

However, the ability of diffusion models to generate highly realistic iris images raises serious security concerns. These synthetic images could potentially be exploited for presentation attack instruments (PAIs), posing a threat to real-world iris recognition systems.

In our study, we leverage the strengths of diffusion models to address the scarcity of publicly available iris datasets while maintaining a focus on both biometric uniqueness and image quality. This approach not only facilitates advancements in iris biometric research but also highlights the importance of ongoing vigilance against potential misuse of this technology.

### E. Our Approach and Motivation

In this work, we aim to address the growing challenges in iris biometric research, including dataset scarcity, privacy concerns, and ethical constraints associated with real iris image collection.

To this end, we use an open-source diffusion framework to generate a database of synthetically created, colored, and highly realistic iris textures. By training on a curated dataset of segmented and noise-reduced iris images, our model produces synthetic irises that are both visually realistic and biometrically unique. Using a Daugman-like iris identification system, we validate the independence of our synthetic samples from the training data through Hamming distance analysis. Additionally, pigmentation analysis confirms that our synthetic dataset captures the full spectrum of natural iris colors.

## II. RELATED WORKS

The growing need for large, diverse iris databases in biometric research has led to various approaches for generating synthetic iris images. These synthetic generation methods can be broadly categorized into three main approaches: texture generation with feature agglomeration [11], patch-based sampling [12-14], and deep learning-based methods, particularly Generative Adversarial Networks (GANs) [32]. Among these approaches, texture generation with feature agglomeration stands out as the only method capable of creating synthetic iris images without requiring real iris images as input, making it particularly valuable for applications where privacy concerns limit access to real data. The other methods, while effective at generating realistic iris patterns, rely on existing iris images for training or reference. Let me examine each of these approaches in detail, as they provide important context for understanding the evolution of synthetic iris generation techniques and their respective advantages and limitations

### A. WVU Synthetic Model Based Dataset and WVU Synthetic Textured Based Dataset [11,15]

There are two WVU synthetic datasets: a "model based" iris dataset and a "texture based" iris dataset [33].

Each iris in the WVU Synthetic Model Based Iris Dataset was generated by creating a hollow 3D cylinder from 500-2500 continuous fibers, then projecting the 3D cylinder into 2D space [11]. The resulting 23 images were strategically brightened and blurred such as to make them closely resemble real irises, both in texture and shape.

Irises in the WVU Textured Model Based Iris Dataset were created through a 2-step process: 1) A Markov Random Field was created to represent the texture of an iris; and 2) Iris features – such as furrows, collarette and crypts – were generated and added to iris generated textures through agglomeration [15]. The result was iris images whose textures and shapes that resembled real irises.

Both datasets contain only greyscale iris images and resembled near infrared image acquisition. Further, 'authentic' subsets were created for both datasets.

Artifacts from the synthetic generation process make iris images from both datasets look as if they were synthetically generated. However, iris patterns were merged with templates containing pupils, sclera, eyelids, and eyelashes.

To verify that the irises in each dataset resembled, a 1D Log Gabor filter bank - like that described by Masek [16]- was used for feature extraction and iris code generation. Finally, the HD comparison distributions for authentic and impostor irises within the synthetic dataset were compared against those of real iris datasets. While this validation demonstrated basic biometric functionality, the synthetic patterns' artificial appearance limited their practical utility for real-world applications and research purposes. The visible artifacts and mechanical regularity in the generated patterns made them easily distinguishable from natural iris textures, highlighting the need for more advanced generation techniques.

### B. CASIA-IrisV4-Syn [12-14]

Currently, numerous studies utilize the CASIA-Iris-V4-Syn dataset for iris recognition [25] and iris privacy [26]. CASIA-Iris-V4-Syn is a database of irises synthetically generated from real irises in the CASIA-Iris-V1 dataset [12, 34]. To prepare the CASIA-Iris-V1 dataset for training, irises were segmented and prepared algorithmically [17]. Then, a patch-based sampling algorithm [13] was used to create new irises from the existing irises. The resulting synthetic iris patterns were merged back into templates of containing pupils, sclera, eyelids, and eyelashes [14].

The generated irises in the CASIA-Iris-V4-Syn dataset closely resembled the real photos in the CASIA-Iris-V1 dataset from which it was created. Two tactics were used to verify the realism of the CASIA-Iris-V4-Syn dataset:

1) A mix of iris experts and lay people were presented real iris photos and synthetic iris photos from the CASIA-Iris-V4-Syn dataset, and asked to determine which irises were real and which irises were fake. It was found that the synthetic iris images were highly realistic.

2.) The authentic and imposter HD distributions of the synthetically generated irises were compared against that of real irises. It is unclear exactly which iris identification system was used; however, the use of HD indicates the use of a 'traditional' Daugman-like process [2].

This method produced much more realistic results, but there was a catch: they needed access to real iris databases to make it work. This became a significant limitation because, as my research shows, getting access to real iris databases is increasingly difficult due to ethical concerns and legal restrictions.

### C. LivDet-Iris-2023-Part1 [18, 19]

The LivDet-Iris-2023-Part1 database was developed as part of the 2023 LivDet-Iris competition, which aimed to advance iris presentation attack detection algorithms [18]. The database included 6,000 synthetically generated Presentation Attack Instrument (PAI) images that adhered to ISO standards, categorized into 'low quality', 'medium quality', and 'high quality' samples. Competition participants were challenged to detect these synthetic images as part of the presentation attack detection task.

The synthetic images were generated using NVIDIA's StyleGAN2-ADA and StyleGAN3 frameworks [19], trained on near-infrared iris scanner data that produced greyscale outputs. However, a significant limitation of this approach was its requirement for an extensive training dataset of over 407,000 real iris images, making it resource-intensive and potentially impractical for many applications.

### D. UND-SFI-2024 [20, 21]

The UND-SFI-2024 dataset was generated using NVIDIA's StyleGAN2-ADA GAN framework, trained on 8,064 ISO-compliant greyscale iris images sourced from three post-mortem iris datasets [20]. The training data included multiple classes of irises, each representing different post-mortem intervals. The project aimed to generate iris images that could simulate the natural deterioration patterns occurring after death, which is crucial for studying potential presentation attacks using deceased persons' irises.

The synthetic images maintained the characteristics of the training data, producing ISO-compliant, greyscale images captured under near-infrared conditions [21]. Both training and generated images included anatomical features and surrounding context, though eyelid and eyelash interference was minimized through specific positioning of the cadavers' eyes. The quality of the generated images was validated against ISO standards for iris identification systems.

However, this approach faced two significant limitations: it required an extensive training dataset of over 8,000 post-mortem iris images, and the framework could only generate greyscale images, restricting its applicability in scenarios requiring color iris representations.

A critical analysis of these methods reveals several persistent limitations. First, most approaches are restricted to grayscale or near-infrared representations, leaving a significant gap in the generation of colored iris images. Second, while some methods achieve visual realism, they often lack comprehensive validation frameworks that simultaneously assess both biometric uniqueness and visual authenticity. Third, the heavy reliance on existing iris images in many approaches raises privacy concerns and perpetuates the challenges of dataset availability.

These limitations underscore the need for a more robust and versatile framework for synthetic iris generation. An ideal solution would need to address multiple criteria simultaneously: generating diverse, realistic colored iris images, ensuring biometric uniqueness, maintaining visual authenticity, and providing comprehensive validation methods. This gap in current methodologies motivates our exploration of diffusion-based approaches, which offer the potential to meet these complex requirements while maintaining rigorous security considerations.

To address these challenges and develop a more comprehensive solution, we began with a carefully curated dataset that would serve as the foundation for our diffusion-based approach.

## III. DATASETS

The development of our synthetic iris generation framework relied on a carefully structured dataset preparation process. Our approach began with a comprehensive collection of high-quality iris images, which underwent extensive preprocessing and refinement to create an optimal training dataset. This chapter details our dataset preparation pipeline, from the initial collection of real iris images to the creation of specialized subsets for model validation. We first describe our starting dataset of manually segmented DSLR-captured images, then outline the rigorous preprocessing steps used to create our training dataset. Additionally, we detail the creation of 'authentic' and 'imposter' subsets, which were crucial for validating the biometric uniqueness of our generated samples. This methodical approach to dataset preparation was essential for ensuring both the quality of our synthetic outputs and the reliability of our validation metrics.

### A. Starting Dataset

We started with an internal set of 6989 DSLR-captured RGB whose irises had been: 1) manually segmented from the sclera, pupil, eyelashes, and eyelids; and 2) rid of noise created by specular reflections.

The dataset includes right and left eyes from the same subjects. All eye images were collected with institutional ethical approval (Indiana University IRB 1409306349).

### B. Training Dataset

Our training set was a subset of the starting dataset following an 11-step quality control process:

1) *Approximating the Iris Pattern's Concentric Limbic and Pupilary Boundaries to The Iris Pattern:* For each image, the pupillary radius was approximated by finding the shortest distance between the center of the image and the closest colored pixel. The limbic radius was approximated by finding the longest distance between the pupillary boundary and a colored pixel in the direction of the edge of the image. Using the estimated pupillary and limbic boundary radii, two concentric circles were imposed over the center of the image. The space in between represented the iris pattern, which was used for further processing.

2) *Discarding Iris Images with Too Little Information:* The 'iris area' – now bound by two concentric circles approximating the limbic and pupillary boundaries – was divided into 12 arch-shaped segments, each of which represented a 30° slice of the iris pattern [27,28]. Any iris image with four or more arch boundaries that did not intersect a portion of colored iris pattern was discarded.

3) *"Unwrapping" the Iris Pattern into Polar Space:* To a) eliminate the black background pixels that were outside of the iris pattern; and b) maximize the efficacy of the imputation of missing iris pattern pixels, each iris pattern was transformed into a rectangular polar coordinate space [29]. Then, a SRCNN model [30] was applied to upscale each image, initially resizing them using bicubic interpolation to the target resolution, which was further refined to 3216x341 pixels through a deep convolutional neural network, thus enhancing image details and quality.

4) *Imputing Missing Parts of The Iris Pattern:* Adobe Photoshop was used to impute missing parts of the unwrapped iris pattern. This was done in a 3-step process: a) selecting all missing pixels of the 3216x341 image; b.) using the selection by 5 pixels in every direction; and c) using Photoshop's "Content Aware" feature in "dissolve" mode to impute missing pixels. Any images that were too dark for Photoshop to sucessfully impute were removed.

5) *White Balance Correction:* To standardize white balance in an image, the mean intensity was first calculated for each pixel across the RGB (Red, Green, Blue) channels. Subsequently, we identified the top one percent of pixels with the highest mean intensity across these color channels. White balance was then standardized based on these pixels. This standardization process adjusts the color temperature of the entire image to match the color temperature derived from these brightest pixels, ensuring consistent color representation throughout the image.

6) *"Wrapping" Iris Pattern Pixels into Concentric Circles:* We wrapped our iris images into concentric rings with uniform dimensions by mapping them back into a Cartesian coordinate system. Although this process introduces noise and artifacts, we chose to re-wrap our iris images prior to training rather than after sampling. This strategy allows the model to generalize the added noise and artifacts, thereby enhancing the robustness of the iris images generated by the model. The resulting images were 1024x1024 squares containing circular and centered irises imposed upon a black background. The pupillary boundary of each iris image was defined by a circle with a radius of approximately 170 pixels from the center of its image. The limbic boundary of each iris image was defined by a circle with a radius of approximately 340 pixels from the center of its image.

7) *Manually Assessing Images for Quality:* Some of the starting images were out of focus and contained noise not removed by the manual noise removal process. As the imputation process often amplified these issues, images of low quality—particularly those affected by low resolution and specular reflections—became more apparent. Consequently, we removed these low-quality images from the training set to ensure better data quality for model training.

8) *Applying Class Labels to The Training Set:* To help prevent the model from creating unrealistic colored irises, training images were assigned color classifications that were strategically reduced to two class labels utilizing OpenAI's CLIP (Contrastive Language-Image Pre-training) model [31]. CLIP excels in interpreting visual concepts through natural language descriptors without requiring task-specific training, offering an efficient solution for our iris classification requirements. The first class encompassed irises where blue-grey and green pixels collectively constituted more than 50% of the total iris area, while the second class comprised irises where the combined proportion of light brown and dark brown pixels exceeded 50% of the iris surface.

9) *Reducing The Size of Training Images:* To optimize the dataset for the diffusion model, each iris image was reduced from a size of 1024x1024 to a size of 256x256 via area interpolation. The pupillary boundary of each downsized iris image was defined by a circle with a radius of approximately 45 pixels from the center of its image. The limbic boundary of each iris image was defined by a circle with a radius of approximately 85 pixels from the center of its image.

10) *Rotating Images of the Training Dataset:* To help the model generalize the data and to reduce the chance of overfitting, each iris image was rotated clockwise 11 times in 30-degree intervals. Effectively, this left 12 images for every image that made it do this point; one image was the original, and 11 were rotations of the original.

11) *Final Training Set Properties:* In sum, 1757 of the original 6989 iris images survived pre-processing. To help the model stick to realistic iris pigmentations, the dataset was divided into two pigment-based classes. Additionally, each of the 1757 images was rotated by 11 times by 30 degrees such as to help the model better generalize the data and to help prevent overfitting. This increased the size of the final training set to 21084 with a size of 256x256, wherein each iris pattern had a pupillary boundary radius of 45 pixels and a limbic boundary radius of 85 pixels.

### C. Defining the 'Authentic' Subset

Our original dataset contained only one image per unique identity. Therefore, we had to make variations of our 1757 training images to create an authentic subset of the training images that could be used to help determine the Hamming distance criterion for our iris identification software.

One portion of our authentic subset were the same 11 rotated versions of each image that was created to help the model to better generalize the data.

Another portion of our 'authentic' images were created by applying 30 generated random noise masks to create 30 'hole punched' versions of each image [Figure 1] which would simulate partial image representations that match the original. Then, hole punches were filled in using Photoshop's 'Content Aware' feature. In total each training image had 41 'authentic' versions of itself that was used to measure our authentic Hamming distance distribution.

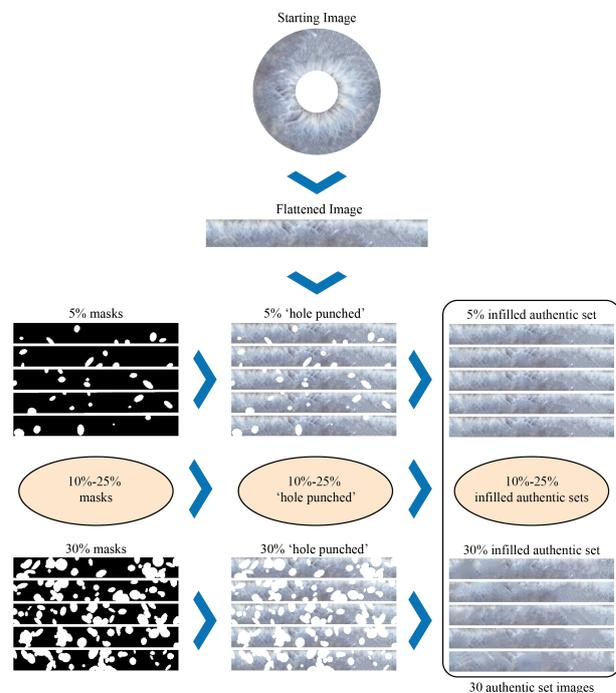

*Figure 1: Overview of process to create 'hole punched' partials of 'authentic' iris images in the training set.*

*D. Defining the 'Imposter' Subset*

For our imposter comparisons, we utilized the training set itself, leveraging all 1,757 unique iris images to create a comprehensive set of imposter comparisons. This approach allowed us to evaluate the distinctiveness of iris patterns across different identities. The imposter Hamming distance distribution was generated by performing an exhaustive comparison, where each image was compared against every other image in the training set, excluding self-comparisons. This resulted in a total of 1,542,646 imposter comparisons, providing a robust statistical basis for establishing our biometric uniqueness threshold.

## IV. APPROACH

*A. Diffusion Network Model*

We leveraged the Guided-Diffusion framework [23], which has demonstrated superior image synthesis capabilities through its innovative architecture and classifier guidance approach. Our implementation utilized a two-class conditional diffusion model trained on a comprehensive dataset comprising of 21,084 iris images, derived from 1,757 unique irises each with 11 rotational augmentations. The model training process extended over 225,000 iterations, with checkpoints systematically saved at 25,000-iteration intervals for progress monitoring and evaluation. [Table 1] details the key hyperparameters employed in our training configuration, which were carefully selected to optimize the model's performance for iris image generation.

| Parameters | Value |
|---|---|
| Image Size | 256 |
| Number of Channels | 128 |
| Number of Residual Blocks | 3 |
| Learn Sigma | True |
| Diffusion Steps | 4000 |
| Noise Schedule | Linear |
| Dropout Rate | 0.3 |
| Class Conditioning | True |
| Learning Rate | 1e-4 |
| Batch Size | 8 |

*Table 1: Parameter configuration for our diffusion model training.*

*B. Iris Biometic Comparison*

To validate the biometric uniqueness of our generated iris images, we developed a custom iris identification process that follows the fundamental principles of traditional iris recognition while incorporating modern computer vision techniques. [Figure 2] illustrates our systematic approach enabling both the validation of our synthetic samples and their comparison against real iris images.

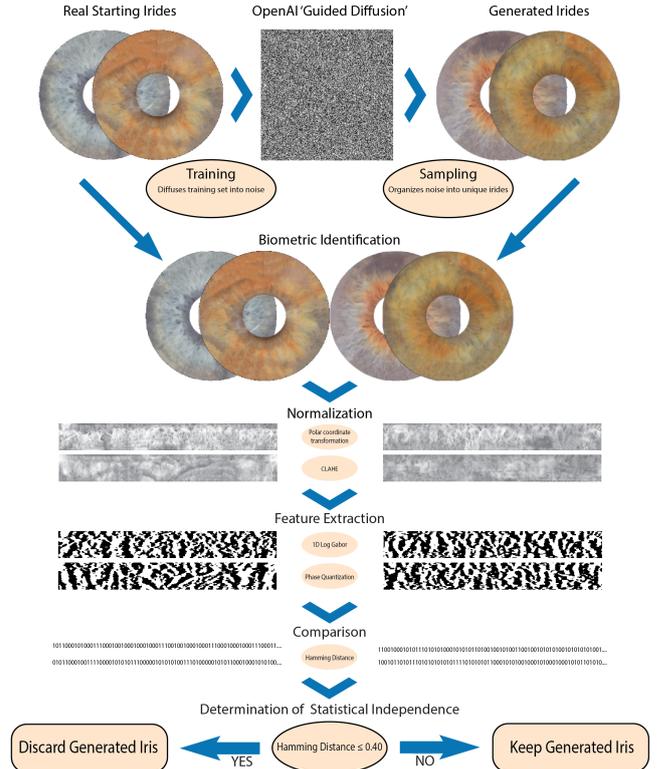

*Figure 2: Overview of our process to filter model output images based on biometrics.*

The following sections detail each component of our identification system and its role in the overall validation process.

1) *Segmentation:* To differentiate between iris patterns and the black backgrounds of their parent images, we use OpenCV to apply Otsu thresholding to grayscale versions of our iris images. Then, we use the OpenCV "contours" function (Suzuki-Abe algorithm) to find the contours of our binary Otsu masks, which was followed by the use of OpenCV's "minEnclosingCircle" function (Wetzl's algorithm) to find the radii of concentric circles enclosing any found contours. Identified concentric circles are sorted by their center coordinates and radii, and filtered based on the known parameters of the input image. If successful, the final result was two concentric circles: one circle representing the pupillary boundary of the iris and another circle representing the limbic boundary of the iris. Segmentation was done in such a way that would fail if pupillary or limbic boundaries radii had more than a three pixels deviation from the training images.

2) *Normalization:* First, images were converted from RGB images into 255 bit greyscale images. Then, a polar coordinate transformation was performed on the pixels between the pupillary and limbic boundaries. The resultant 'unwrapped' iris images were 45 pixels in the radial dimension and 360 pixels in the angular dimension. Finally, contrast limited adapted histogram equalization [2] was applied to each image.

3) *Feature Exraction:* To extract iris features from our normalized iris images, we utilized Masek's 1D Log Gabor method [11] using a single filter with a wavelength of 18 pixels and a sigma/f of 0.5. We picked Masek's approach because of its robustness to rotational invariance, which is a trait we highly valued considering: 1) the use of rotated images in the training set; and 2) the tendency of diffusion networks to rotate images.

The resultant filter response for each image was a 45x360x2 complex-valued matrix. Each of the complex-valued filter responses was quantized into two bits, depending on its phase. Ultimately, each image was condensed into a standardized 32400-bit iris code template.

*4) Comparison of Iris Code Templates:* Hamming distance was used to compare iris code templates. The higher the hamming distance is between two iris code templates, the more likely the they are to be statistically independent from each other.

To ensure that iris code template comparisons were rotationally invariant, we utilized circular bit shifts along the angular axis of iris code templates. Each bit shift corresponded to a 1 degree rotation of the parent iris image. In total, 360 separate hamming distances were computed for each iris pair comparison – with each hamming distance accounting for 1 additional degree of rotation between the two irises until an entire phase was completed. The lowest hamming distances calculated between two iris code templates was retained.

*5) Determining Our Hamming Distance Cutoff Critereon*

To determine the appropriate Hamming distance cutoff criterion for our iris biometric comparison approach, we first compared authentic and impostor distributions of Hamming distances. We calculated these distributions by:

1. Authentic comparisons: comparing each of the 1,757 original images against its 41 variations (11 rotated versions and 30 hole-punched versions)
2. Impostor comparisons: comparing each of the 1,757 original images against all other original images in the training set, excluding comparisons with itself

*C. Iris Color Comparison*

To validate that our generated iris images encompass the full spectrum of natural iris variation, we implemented a rigorous color analysis framework using an SVM-based Iris Color Quantification Tool [22]. This tool was applied to both our training set (21,084 iris images, derived from 1,757 unique irises) and generated dataset (17,695 images), mapping each pixel to a comprehensive colors palette that we mentioned before. This diverse color palette was specifically chosen to capture the subtle variations present in natural iris pigmentation.

For both sets, we applied an isometric log-ratio (ILR) transformation followed by principal component analysis (PCA) to the color quantification data to visualize both color space distributions and overlap.

## V. RESULTS & DISCUSSION

### A. Checking the Model for Signs of Overfitting

2048 samples were created with each checkpoint. For comprehensive evaluation, we conducted four types of comparisons between the generated images and the original dataset (1757 images):

1. FID scores for all generated samples (Original sampling images FID)
2. FID scores for CLIP-filtered samples (Filtered sampling images FID), where the filtering removed samples with CLIP similarity scores below 0.95 based on text descriptions of noisy and common iris images [Table 2] presents these evaluation metrics across different training iterations from 25k to 250k. Both the original and filtered FID scores demonstrated consistent improvement throughout the training process, decreasing from 3.00 at 25k iterations to 0.154 for original samples and 0.148 for filtered samples at 250k iterations. Notably, the classified sampling strategy consistently achieved lower FID scores, demonstrating superior generation quality compared to the original sampling method. Through the CLIP model, we effectively filtered out generated images with quality issues such as overexposure, underexposure, unclear colors, and visible noise. The rightmost column in Table 2 shows the number of images that passed this CLIP-based quality filtering at each checkpoint, with the count stabilizing around 1,900 samples per checkpoint after 125k iterations.

| Iteration | Original sampling images FID | Filtered sampling images FID | Number of classified sampling images |
|---|---|---|---|
| 25k | 3.00 | - | 0 |
| 50k | 0.319 | 0.295 | 1778 |
| 75k | 0.269 | 0.231 | 1718 |
| 100k | 0.235 | 0.214 | 1883 |
| 125k | 0.207 | 0.190 | 1902 |
| 150k | 0.191 | 0.177 | 1907 |
| 175k | 0.184 | 0.165 | 1893 |
| 200k | 0.181 | 0.162 | 1907 |
| 225k | 0.162 | 0.152 | 1992 |
| 250k | 0.154 | 0.148 | 1976 |

*Table 2: Evaluation metrics across diffusion model checkpoints.*

Based on these evaluation results showing optimal FID scores at 250k iterations, we selected this checkpoint for our final sample generation. After applying our comprehensive CLIP-based quality control process to a larger generation batch, we obtained a final dataset of 17,695 high-quality images. Representative samples of the filtered results are illustrated in [Figure 3], demonstrating the high quality and diversity of our generated iris images generate a wide range of realistic iris colors, from blue-green to dark brown. The six examples shown (from left to right) include three variations of grey-blue irises, green, light brown, and dark brown, reflecting the natural color distribution observed in human irises.

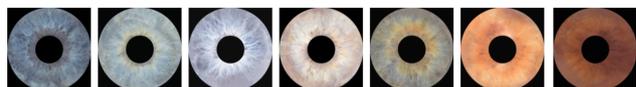

*Figure 3: Representative synthetic iris images generated at 250,000 iterations. The samples demonstrate the model's ability to*

### B. Biometric Validation of Iris Images

We evaluated the biometric uniqueness of our synthetic iris patterns by analyzing HD using our iris identification *approach*. The kernel density estimation plot of HD distribution from comparing our 17,695 synthetic samples against the imposter dataset (the 1,757 training images) [Figure 4] shows a mean of $0.4622 \pm 0.0070$ (±1 SD).

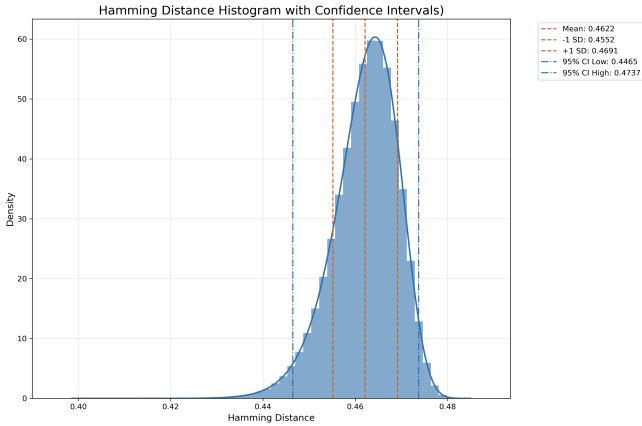

*Figure 4 : Kernel density estimation plots of Hamming distance distributions comparisons between synthetic samples and the training set*

By analyzing 100 potential threshold values between 0 and 1, we found that a Hamming distance threshold that provided optimal separation between authentic and impostor distributions (Figure 5). At this 0.4 threshold, we observed a false acceptance rate (FAR) of $3 * 10^{-6}$ across our test comparisons. Using this criterion, iris code templates with a Hamming distance less than the threshold are considered to come from the same iris, while those with larger distances are considered to come from different irises.

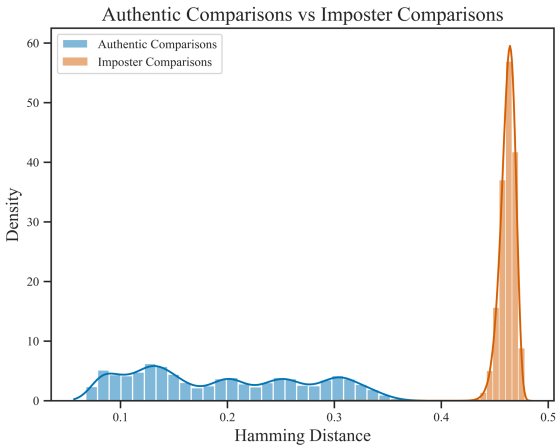

*Figure 5: Combined Histograms of Authentic and Imposter Hamming Distance Comparisons.*

This illustrates that our diffusion model successfully generated unique iris patterns while maintaining the characteristic features of real iris images.

### C. Color Validation of Iris Images

To evaluate the pigmentation distribution of our synthetic iris images, we performed comparative analysis between our 17,695 generated samples and 1,757 training images. The PCA results [Figure 8] provide a comprehensive view of the pigmentation distributions: (a) shows the relationship between training samples (blue points) and synthetic samples (orange points), revealing substantial overlap and complementary coverage patterns in the principal component.

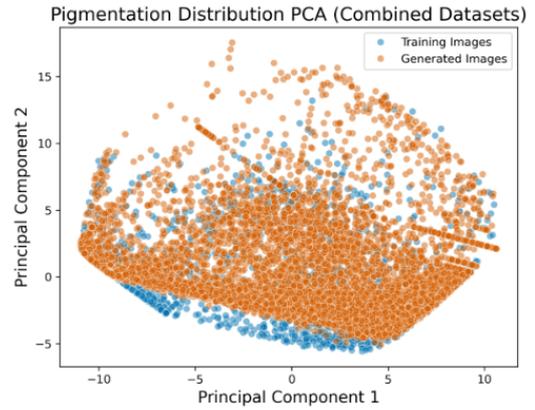

(a)

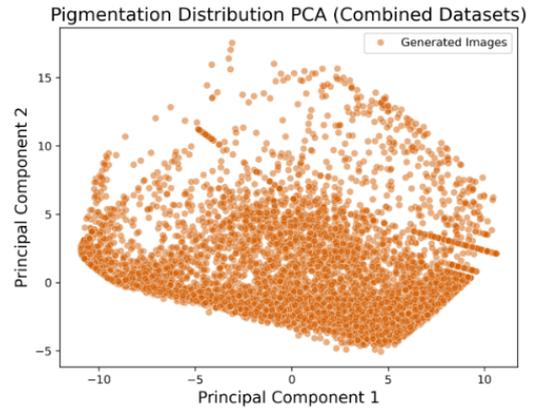

(b)

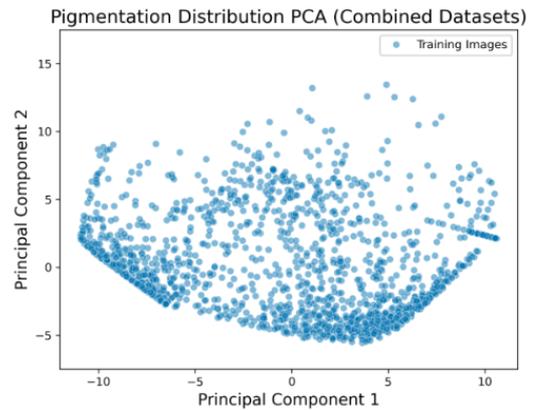

(c)

*Figure 6: Principal Component Analysis (PCA) visualization of iris pigmentation distributions. (a) Combined distribution showing the overlap between training samples (blue points) and synthetic samples (orange points) in the principal component space. (b) Distribution of synthetic samples demonstrating broad coverage of the pigmentation space. (c) Distribution of training showing the natural pigmentation patterns.*

space, (b) illustrates the full distribution of our synthetic samples across the pigmentation space, and (c) displays the training data distribution. These visualizations demonstrate that our synthetic samples effectively capture and extend the natural color distribution present in the training data, with appropriate coverage across the principal component space while maintaining realistic pigmentation patterns that align with biological expectations.

Our validation methodology employed a two-fold Euclidean distance analysis in the quantized color space. The intra-dataset analysis, illustrated in [Figure 6], examined the color distribution within the training set, revealing a characteristic bimodal distribution with peaks at distances of 0.0-0.2 and 0.6-0.9. This bimodal pattern reflects the natural clustering of human iris colors while maintaining continuous variation between major color groups. The inter-dataset comparison, shown in [Figure 7], evaluated the minimum Euclidean distance between each generated image and the training set, demonstrating that our generated images maintain close alignment with natural color distributions while exhibiting appropriate levels of variation.

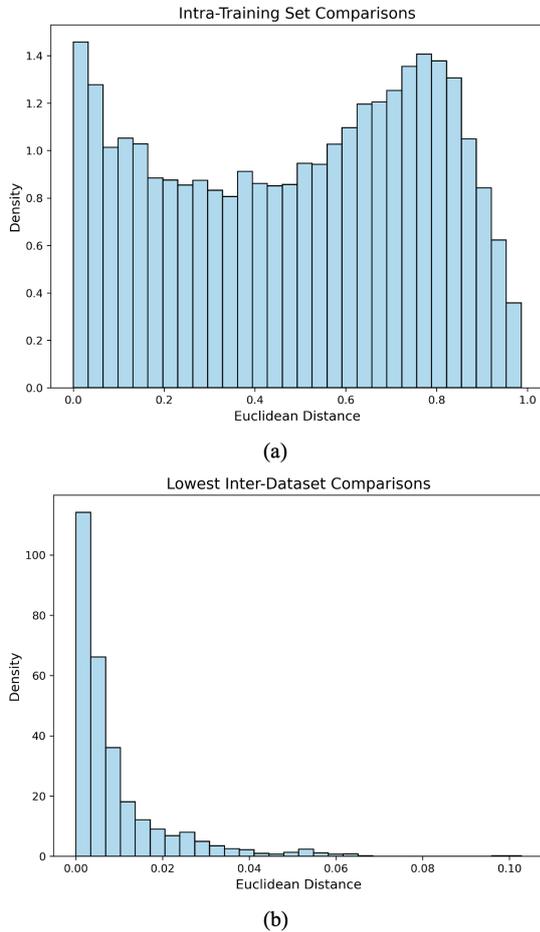

(a)

(b)

*Figure 7: Distribution analysis of iris color patterns. (a) Intra-training set comparison showing the Euclidean distance distribution between pairs of training images. (b) Inter-dataset comparison showing the minimum Euclidean distances between generated images and their closest matches in the training set.*

### D. Merging Sampled Irises Into Realistic Templates

To help contextualize the generated iris images within the greater context of the eye, scripts were 1) merge generated irises with a standardized template of anatomical features such as the scalara, eyelids, eyelashes, and skin and 2) add shadows and blending to make the merged images look more natural [Figure 8], thus completing the biometrically unique and realistic Iris Database made available with this work.

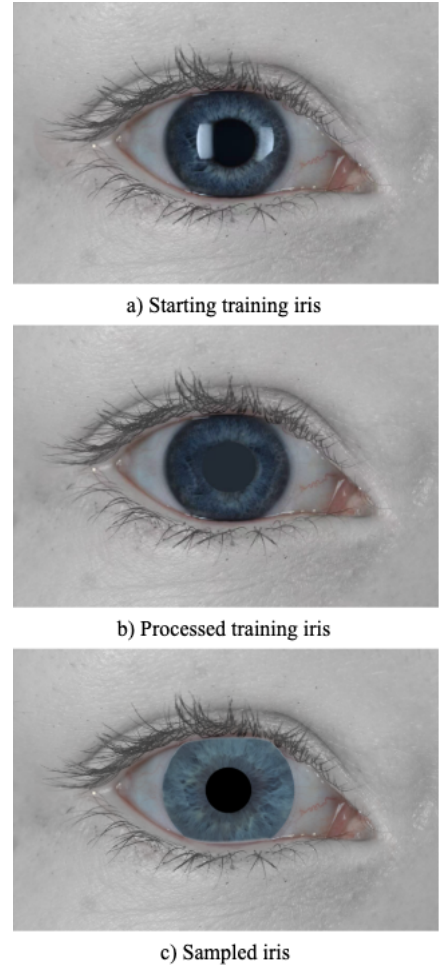

a) Starting training iris

b) Processed training iris

c) Sampled iris

*Figure 8: Starting real eye, processed training eye with pupil removed and merged back into template; and generated diffusion eye merged back into template.*

## VI. Conclusion

Our study demonstrates the effectiveness of diffusion networks in generating synthetic iris images that meet both biometric and aesthetic requirements. By training on a carefully preprocessed dataset of 1,757 real iris images, our model successfully generated 17,695 high-quality synthetic iris patterns that are biometrically unique, as evidenced by Hamming distance distributions centered well above threshold. The FID scores improved consistently throughout training, reaching optimal performance at 250,000 iterations, with filtered samples achieving an FID of 0.148, indicating high-quality generation. Our iris identification approach combines classical recognition principles with modern computer vision techniques to create a reliable validation framework.

However, several critical areas warrant further investigation and improvement:

1. Model Efficiency Enhancement:
   - Optimizing the training process to reduce the computational resources required, particularly given the current requirement of 250,000 iterations

- Investigating methods to accelerate the sampling process while maintaining generation quality
- Exploring techniques to reduce the preprocessing complexity while preserving data quality
2. Security Considerations:
- Evaluating the potential vulnerabilities of diffusion-generated iris images to presentation attacks
- Developing robust detection methods to identify synthetic iris images in biometric systems
- Investigating the integration of watermarking or other tracking mechanisms to prevent misuse of synthetic images
3. Quality and Diversity Enhancement:
- Investigating methods to expand the diversity of generated samples while maintaining realism
- Developing more sophisticated quality assessment metrics specifically for synthetic iris images
- Improving the representation of rare iris patterns and colors in the generated samples
- Developing guidelines for the responsible use of synthetic iris images in research and development

These findings suggest that while diffusion networks offer a promising solution for creating synthetic iris databases, continued research is needed to address efficiency, security, and practical implementation challenges. Future work should focus on balancing the trade-offs between generation quality, computational efficiency, and security considerations while maintaining the biometric utility of the generated images.

The development of more efficient architectures, improved security measures, and comprehensive validation frameworks will be crucial for the widespread adoption of diffusion-based synthetic iris generation in both research and practical applications. Additionally, establishing clear guidelines and best practices for the use of synthetic iris data will be essential for ensuring responsible development in this field.

## VII. ACKNOWLEDGEMENTS

This work was supported in part by the U.S. Department of Justice under Grant 15PNIJ-23-GG-04206-RESS. We thank all the volunteers that consented to this research.